\documentclass[11pt]{article}
\date{}
\usepackage{UF_FRED_paper_style}
\usepackage{lipsum}  
\usepackage{float}
\usepackage{hyperref}
\usepackage{longtable}
\usepackage[table]{xcolor} 
\title{Examining the Robustness of Large Language Models across Language Complexity
}

\author{Jiayi Zhang\\
    University of Pennsylvania\\
    \href{mailto:joycez@upenn.edu}{\texttt{joycez@upenn.edu}}
}

\begin{document}

{\setstretch{.8}

\maketitle

% %%%%%%%%%%%%%%%%%%

\begin{abstract}

\noindent
With the advancement of large language models (LLMs), an increasing number of student models have leveraged LLMs to analyze textual artifacts generated by students to understand and evaluate their learning. These student models typically employ pre-trained LLMs to vectorize text inputs into embeddings and then use the embeddings to train models to detect the presence or absence of a construct of interest. However, how reliable and robust are these models at processing language with different levels of complexity? In the context of learning where students may have different language backgrounds with various levels of writing skills, it is critical to examine the robustness of such models to ensure that these models work equally well for text with varying levels of language complexity. Coincidentally, a few (but limited) research studies show that the use of language can indeed impact the performance of LLMs. As such, in the current study, we examined the robustness of several LLM-based student models that detect student self-regulated learning (SRL) in math problem-solving. Specifically, we compared how the performance of these models vary using texts with high and low lexical, syntactic, and semantic complexity measured by three linguistic measures.

\vspace{1em}

\noindent \textbf{Keywords:} Large Language Models; Model Robustness; Language Complexity; Self-Regulated Learning

\end{abstract}
}

% %%%%%%%%%%%%%%%%%%%%%%%%%%%%%%%%%%%%%%%%%%%%%%%%%%%%%%%%%%
% %%%%%%%%%%%%%%%%%%%%%%%%%%%%%%%%%%%%%%%%%%%%%%%%%%%%%%%%%%
% BODY OF THE DOCUMENT
% %%%%%%%%%%%%%%%%%%%%%%%%%%%%%%%%%%%%%%%%%%%%%%%%%%%%%%%%%%
% %%%%%%%%%%%%%%%%%%%%%%%%%%%%%%%%%%%%%%%%%%%%%%%%%%%%%%%%%%

\section{Introduction}
The advent of large language models (LLMs) has created an opportunity in the field of education to develop automated detectors that analyze students' textual artifacts. These detectors typically employ pre-trained LLMs to vectorize text inputs into embeddings, which are then used as features to train models that detect the presence or absence of a construct of interest. Previous studies have used this approach to examine students' use of self-regulated learning during problem-solving \citep{zhang2024srl_llms} and to evaluate the quality and attributes of peer feedback \citep{darvishi2022trustworthy}.

However, how reliable and robust are these models at processing language with different levels of complexity? In the field of education where students may have different language backgrounds and varying levels of writing skills, model with unreliable performance may potentially create unfair evaluation against certain groups of students \citep{sha2021fairness}. It is critical to examine the robustness of such models to ensure they work equally well for texts with varying levels of complexity. Coincidentally, a few (but limited) research shows that the complexity of language can indeed impact the performance of LLMs. For example, \citep{dentella2023ai_performance} found that low-frequency aspects of language (e.g., semantic anomalies, complex nested hierarchies, and self-embeddings) reveal limitations in the ability of GPT-3 to understand the subtleties of language. Additionally, language models often give inaccurate answers to questions with ungrammatical prompts that involve different language phenomena (e.g., anaphora, center embedding, comparative sentences) \citep{dentella2023language_models}.

Given these concerns, the current study examined the robustness of several existing detectors which leverage LLMs to detect student self-regulated learning (SRL) in math problem-solving. Specifically, we compared how the performance of these detectors differs with texts that have high and low lexical, syntactic, and semantic complexity, measured by three linguistic measures, respectively.

\section{Methods}
In this section, we first describe the learning platform and the sample data (see Section ~\ref{subsec:platform}). Next, we summarize the development of several machine learning (ML) models that are trained to automatically detect students' use of self-regulated learning (SRL) in math problem-solving (see Section ~\ref{subsec:detectors}). Finally, we describe the analysis that evaluates the robustness of these models based on language complexity (see Section ~\ref{subsec:complexity}).

\subsection{Learning Platform and Data}\label{subsec:platform}
\begin{figure}[H]
    \centering
        \includegraphics[scale=.4]{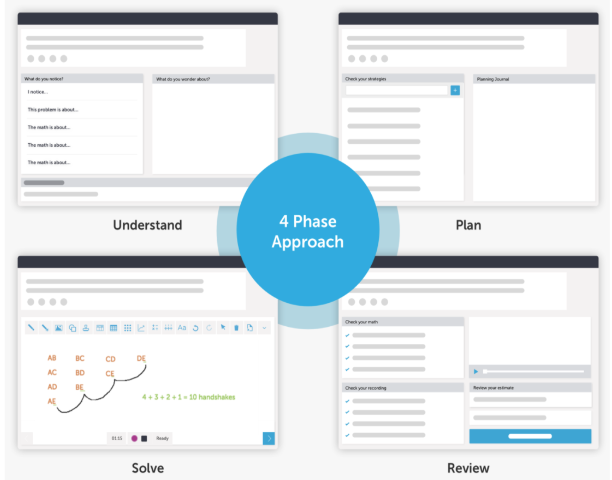}
    \caption{Screenshots of CueThink’s Four Phase Approach.}
    \label{fig:cuethink}
\end{figure}

CueThink is a digital learning application that focuses on enhancing middle school students' math problem-solving skills by encouraging them to engage in self-regulated learning and to develop math language to communicate problem-solving processes. As ~\ref{fig:cuethink} shows, CueThink structures a math problem into a Thinklet, a process that includes four phases: Understand, Plan, Solve, and Review. Within each phase, both open-ended and multiple-choice questions are asked to scaffold students' problem-solving processes. Specifically, in the Understand phase, students are asked what they notice and what they wonder, as these questions encourage students to actively look for information in the problem, creating a problem representation. In the Plan phase, students build on what they have established in the Understand phase and develop a plan, outlining how they will solve the problem. In the Solve phase, students present and explain their solutions. In the Review phase, students give the final answer to the math problem and reflect on whether the answer makes sense and whether their communication is clear.

In total, we collected students' textual responses from 182 Thinklets, which were created by 79 students working on 24 unique math problems. A full description of the platform and student demographics can be found in \citep{zhang2022using}.

\subsection{Developing SRL Detectors}\label{subsec:detectors}

\subsubsection{Defining and Coding SRL Constructs}

Developing automated detectors of SRL begins with operationalizing SRL constructs and manually labeling the presence or absence of these constructs in the data. In this study, by analyzing students' textual responses within each Thinklet, we first identified five SRL constructs that are salient in the dataset. These five constructs are: 1) numerical representation (NR), 2) contextual representation (CR), 3) strategy orientation (SO), 4) outcome orientation (OO), and 5) data transformation (DT). The operationalization of these constructs is grounded in Winne’s SMART model (an acronym for searching, monitoring, assembling, rehearsing, and translating), which outlines the five cognitive operations involved in SRL processes \citep{winne2017learning_analytics}.

As shown in Table \ref{tab:srl}, these five constructs fall under the assembling and translating categories in the SMART model. They reflect how students assemble information when creating a problem representation (numerical representation and contextual representation), how they assemble information in forming a plan (outcome orientation and strategy orientation), and whether they manipulate the ways information is represented.
Using the definitions outlined in Table  \ref{tab:srl}, two researchers first coded the same set of Thinklets to establish reliability (\(\kappa_{\text{NR}}\) = 0.83, \(\kappa_{\text{CR}}\) =0.63, \(\kappa_{\text{SO}}\) = 0.74, \( \kappa_{\text{OO}}\)  = 0.78,\(\kappa_{\text{DT}}\) = 0.74). Once an acceptable reliability had been reached, the two researchers coded individually for the remaining Thinklets. These labels were then used as the ground truth to train machine learning models.

\begin{table}[H]
\centering
\renewcommand{\arraystretch}{1.2}  % Increase row spacing for readability
\caption{SMART Category, SRL Indicator, and Definition}
\resizebox{\textwidth}{!}{  % Resizes table to fit within text width
\begin{tabular}{l l p{10cm}} \hline
\textbf{SMART Category} & \textbf{SRL Indicator} & \textbf{Working Definition} \\ \hline
\multirow{4}{*}{Assembling} 
    & Numerical Representation (NR) & The learner’s representation of the problems includes numerical components and demonstrates a level of understanding of how the numerical values are used in the math problem. \\
    & Contextual Representation (CR) & The learner’s representation of the problem includes contextual details relating to the setting/characters/situations within the given math problem. \\
    & Strategy Orientation (SO) & Learners explicitly state a plan for how they will find the answer for the given math problem, decomposing information into a step-by-step process. \\
    & Outcome Orientation (OO) & The learner provides only a numerical estimate of the final answer for the given math problem, suggesting that learners are focused on the output instead of the process itself. \\ \hline
Translating & Data Transformation (DT) & The learner manipulates the ways information is represented to them in the problem to find a solution. This suggests active problem solving. \\ \hline
\end{tabular}
}
\label{tab:srl}
\end{table}

\subsubsection{Training LLM-based SRL detectors}
To train models that automatically detect the presence or absence of the SRL constructs, we first concatenated all the textual responses within a Thinklet. The concatenated text was then vectorized using OpenAI’s sentence-embedding-3-short model \citep{neelakantan2022text_code_embeddings}, which generates a 1536-dimensional embedding based on each word in a sentence and the words surrounding it. This step converts the text-based input into a numerical representation as output. Using the numerical outputs and the manually coded labels, we trained a neural network model with one hidden layer to predict the presence or absence of each SRL construct.

\subsection{Language Complexity Measures}\label{subsec:complexity}
To examine the robustness of these models based on language complexity, we selected three linguistic measures that assess the lexical, syntactic, and semantic complexity of the text, respectively. Specifically, for the text in each Thinklet, we computed the Mass score \citep{torruella2013lexical_statistics} using Equation \ref{eq:mass}. The Mass score measures lexical complexity by examining the relationship between the number of different words (types) and the total number of words (tokens), while also considering sentence length (tokens). This measure is preferred over other lexical complexity measures, such as the type-token ratio, which does not take sentence length into consideration and could potentially confound the assessment. For sentences of the same length, a higher Mass score indicates a more diverse use of words, signifying higher lexical complexity.

\begin{equation} \label{eq:mass}
\text{Mass} = \frac{\log(\text{types}) - \log(\text{tokens})}{\log_2(\text{tokens})}
\end{equation}

Additionally, we computed the syntactic simplicity value using Coh-Metrix \citep{mcnamara2010cohmetrix}. Syntactic simplicity measures how syntax is structured by the number of words and clauses in a sentence, or the number of words before the main verb in the sentence. When the syntax is more complex, readers may have more difficulty creating a coherent understanding of the sentence’s meaning, which can potentially hinder an LLM’s ability to process it. The syntactic simplicity value inversely reflects the syntactic complexity of a text, meaning the text has lower syntactic complexity when the simplicity value is high. We also computed the deep cohesion measure from Coh-Metrix to reflect the semantic complexity of a text. The deep cohesion reflects how events and ideas are related throughout the entire text, with greater overlap suggesting greater overall cohesion \citep{mcnamara2010cohmetrix}.

For each linguistic measure, a median value was identified (see Table \ref{tab:descriptive}). Using this median, we classified all Thinklets into two groups: one group containing text with complexity higher than the median, and the other group containing text with complexity value equal to or lower than the median. We then evaluated and compared the performance of the SRL detectors between the two complexity groups (high vs. low) for each linguistic measure.

\begin{table}[h]
    \centering
    \caption{Descriptive Statistics of the Three Linguistic Measures}
    \begin{tabular}{lcccccc}
        \hline
        & Range & Min. & Median & Max & Mean & Standard Deviation \\
        \hline
        MASS & 0-1 & 0 & 0.03 & 0.34 & 0.03 & 0.03 \\
        Syntactic Simplicity & 0-100 & 2 & 76 & 100 & 69 & 26 \\
        Deep Cohesion & 0-100 & 1 & 70 & 100 & 56 & 40 \\
        \hline
    \end{tabular}
    \label{tab:descriptive}
\end{table}

\section{Results}
\subsection{Model Performance}
We first evaluated the model performance by using 10-fold student-level cross-validation to ensure the general success of these models in detecting the SRL constructs. The average and the standard deviation (SD) of Area Under the Receiver Operating Characteristic curve (AUC) was computed for each model. The average AUC is 0.918 (SD=0.075) for numerical representation, 0.801 (SD=0.141) for contextual representation, 0.772 (SD=0.145) for outcome orientation, and 0.783 (SD=0.109) for data transformation. These results indicate that the models are generally successful at capturing the four SRL behaviors in students’ textual responses during the problem-solving process. Due to the rarity of strategy orientation (only 14 Thinklets were labeled with this construct), a detector could not be built for this construct and therefore excluded from the following analysis. 

\subsection{Model Robustness by Language Complexity}
To examine the robustness of these models based on language complexity, we compared how the models performed for text with high and low complexity based on the three linguistic measures (i.e., MASS, syntactic simplicity, and deep cohesion). As shown in Table \ref{tab:model_performance}, for each linguistic measure, we computed and compared the performance (i.e., AUCs) of each SRL detector between the two  complexity groups (high vs. low). We highlight the comparisons where the difference is noticeable between the two groups, indicated by an absolute difference in AUCs greater than one.

As shown in Table \ref{tab:model_performance}, all four SRL models demonstrate a comparable performance when processing text with low and high lexical complexity. For syntactic complexity, we find when predicting contextual representation (CR), the model performs better when the text is more syntactically complex (low in syntactic simplicity score). However, the model that predicts data transformation (DT) is better when the text is less syntactically complex. For semantic complexity, measured by deep cohesion, we find that the model that predicts numerical representation (NR) performs better when the text is high in cohesion, whereas the model that predicts contextual representation (CR) performs better when the text is low in cohesion.

\begin{table}[h]
    \centering
    \caption{Model Performance by the Three Linguistic Measures}
    \begin{tabular}{lcc|cc|cc}
        \hline
        & \multicolumn{2}{c|}{MASS} & \multicolumn{2}{c|}{Syntactic Simplicity} & \multicolumn{2}{c}{Deep Cohesion} \\
        \hline
        & Low & High & Low & High & Low & High \\
        \hline
        NR & 0.876 & 0.926 & 0.915 & 0.912 & \cellcolor{yellow}0.845 & \cellcolor{yellow} 0.955 \\
        CR & 0.710 & 0.799 & \cellcolor{yellow} 0.851 & \cellcolor{yellow} 0.739 & \cellcolor{yellow} 0.813 & \cellcolor{yellow} 0.652 \\
        OO & 0.721 & 0.758 & 0.725 & 0.772 & 0.747 & 0.707 \\
        DT & 0.718 & 0.754 & \cellcolor{yellow} 0.683 & \cellcolor{yellow} 0.815 & 0.690 & 0.706 \\
        \hline
    \end{tabular}
    \label{tab:model_performance}
\end{table}

\section{Discussion and Conclusion}
In this study, we examined the robustness of several LLM-based detectors with respect to the complexity of the input text. We aimed to evaluate whether these detectors perform consistently across text with different levels of lexical, syntactic, and semantic complexity. Through the analysis, we found that during the math problem-solving process, students' textual responses are more homogeneous in lexical complexity than in syntactic and semantic complexity. The small variance in lexical complexity may explain why the models had comparable performance when processing text with high and low lexical complexity. For syntactic complexity, we observed that when predicting contextual representation, where the model identifies behaviors related to how students assemble contextual details in the problem, the model performs better when the text is more syntactically complex. However, when predicting data transformation, the model performs better on text with simpler syntax. The opposite relationships between model performance and syntactic complexity in these two cases may be due in part to the different ways the two SRL behaviors are represented in textual responses. Specifically, compared to representing a problem space contextually (contextual representation), data transformation identifies cases where students convert how information is presented, which normally involves students converting the relationship observed in the problem into a mathematical equation. The formulation of an equation and the words surrounding it is typically simpler in syntax. Thus, the model might work better for text with simpler syntax when detecting data transformation.

Additionally, we found that the model performs better in predicting numerical representation when the text is more cohesive, while it performs better in predicting contextual representation when the text is less cohesive. This may be because numerical representation requires students to demonstrate an understanding of how numerical values are used in the math problem, which benefits from cohesive text. In contrast, contextual representation involves assembling contextual details within the problem without necessarily showing how information across sentences is connected.
To the best of our knowledge, no previous studies have examined the relationship between model performance and language complexity in assessing the robustness of student models. We plan to replicate this analysis with other datasets, potentially involving texts with a broader range of complexity. As LLMs are increasingly used in student models with the potential to inform pedagogical decisions that affect student learning, there is an urgent need for analyses that examine the robustness and fairness of these models to ensure consistent performance across all students and their textual artifacts.

% %%%%%%%%%%%%%%%%%%%%%%%%%%%%%%%%%%%%%%%%%%%%%%%%%%%%%%%%%%
% %%%%%%%%%%%%%%%%%%%%%%%%%%%%%%%%%%%%%%%%%%%%%%%%%%%%%%%%%%
% REFERENCES SECTION
% %%%%%%%%%%%%%%%%%%%%%%%%%%%%%%%%%%%%%%%%%%%%%%%%%%%%%%%%%%
% %%%%%%%%%%%%%%%%%%%%%%%%%%%%%%%%%%%%%%%%%%%%%%%%%%%%%%%%%%
\medskip

\bibliography{references.bib} 

\begin{thebibliography}{}

\bibitem [\protect \citeauthoryear {%
Darvishi%
, Khosravi%
, Sadiq%
\BCBL {}\ \BBA {} Gašević%
}{%
Darvishi%
\ \protect \BOthers {.}}{%
{\protect \APACyear {2022}}%
}]{%
darvishi2022trustworthy}
\APACinsertmetastar {%
darvishi2022trustworthy}%
\begin{APACrefauthors}%
Darvishi, A.%
, Khosravi, H.%
, Sadiq, S.%
\BCBL {}\ \BBA {} Gašević, D.%
\end{APACrefauthors}%
\unskip\
\newblock
\APACrefYearMonthDay{2022}{}{}.
\newblock
{\BBOQ}\APACrefatitle {Incorporating AI and learning analytics to build trustworthy peer assessment systems} {Incorporating ai and learning analytics to build trustworthy peer assessment systems}.{\BBCQ}
\newblock
\APACjournalVolNumPages{British Journal of Educational Technology}{53}{4}{844--875}.
\newblock
\begin{APACrefDOI} \doi{10.1111/bjet.13233} \end{APACrefDOI}
\PrintBackRefs{\CurrentBib}

\bibitem [\protect \citeauthoryear {%
Dentella%
, Günther%
\BCBL {}\ \BBA {} Leivada%
}{%
Dentella%
, Günther%
\BCBL {}\ \BBA {} Leivada%
}{%
{\protect \APACyear {2023}}%
}]{%
dentella2023language_models}
\APACinsertmetastar {%
dentella2023language_models}%
\begin{APACrefauthors}%
Dentella, V.%
, Günther, F.%
\BCBL {}\ \BBA {} Leivada, E.%
\end{APACrefauthors}%
\unskip\
\newblock
\APACrefYearMonthDay{2023}{}{}.
\newblock
{\BBOQ}\APACrefatitle {Systematic testing of three Language Models reveals low language accuracy, absence of response stability, and a yes-response bias} {Systematic testing of three language models reveals low language accuracy, absence of response stability, and a yes-response bias}.{\BBCQ}
\newblock
\APACjournalVolNumPages{Proceedings of the National Academy of Sciences}{120}{51}{e2309583120}.
\newblock
\begin{APACrefDOI} \doi{10.1073/pnas.2309583120} \end{APACrefDOI}
\PrintBackRefs{\CurrentBib}

\bibitem [\protect \citeauthoryear {%
Dentella%
, Murphy%
, Marcus%
\BCBL {}\ \BBA {} Leivada%
}{%
Dentella%
, Murphy%
\BCBL {}\ \protect \BOthers {.}}{%
{\protect \APACyear {2023}}%
}]{%
dentella2023ai_performance}
\APACinsertmetastar {%
dentella2023ai_performance}%
\begin{APACrefauthors}%
Dentella, V.%
, Murphy, E.%
, Marcus, G.%
\BCBL {}\ \BBA {} Leivada, E.%
\end{APACrefauthors}%
\unskip\
\newblock
\APACrefYearMonthDay{2023}{}{}.
\newblock
\APACrefbtitle {Testing AI performance on less frequent aspects of language reveals insensitivity to underlying meaning.} {Testing ai performance on less frequent aspects of language reveals insensitivity to underlying meaning.}
\PrintBackRefs{\CurrentBib}

\bibitem [\protect \citeauthoryear {%
McNamara%
, Louwerse%
, McCarthy%
\BCBL {}\ \BBA {} Graesser%
}{%
McNamara%
\ \protect \BOthers {.}}{%
{\protect \APACyear {2010}}%
}]{%
mcnamara2010cohmetrix}
\APACinsertmetastar {%
mcnamara2010cohmetrix}%
\begin{APACrefauthors}%
McNamara, D\BPBI S.%
, Louwerse, M\BPBI M.%
, McCarthy, P\BPBI M.%
\BCBL {}\ \BBA {} Graesser, A\BPBI C.%
\end{APACrefauthors}%
\unskip\
\newblock
\APACrefYearMonthDay{2010}{}{}.
\newblock
{\BBOQ}\APACrefatitle {Coh-Metrix: Capturing Linguistic Features of Cohesion} {Coh-metrix: Capturing linguistic features of cohesion}.{\BBCQ}
\newblock
\APACjournalVolNumPages{Discourse Processes}{47}{4}{292--330}.
\newblock
\begin{APACrefDOI} \doi{10.1080/01638530902959943} \end{APACrefDOI}
\PrintBackRefs{\CurrentBib}

\bibitem [\protect \citeauthoryear {%
Neelakantan%
\ \protect \BOthers {.}}{%
Neelakantan%
\ \protect \BOthers {.}}{%
{\protect \APACyear {2022}}%
}]{%
neelakantan2022text_code_embeddings}
\APACinsertmetastar {%
neelakantan2022text_code_embeddings}%
\begin{APACrefauthors}%
Neelakantan, A.%
, Xu, T.%
, Puri, R.%
, Radford, A.%
, Han, J\BPBI M.%
, Tworek, J.%
\BDBL {}Weng, L.%
\end{APACrefauthors}%
\unskip\
\newblock
\APACrefYearMonthDay{2022}{}{}.
\newblock
\APACrefbtitle {Text and Code Embeddings by Contrastive Pre-Training.} {Text and code embeddings by contrastive pre-training.}
\newblock
\begin{APACrefURL} \url{http://arxiv.org/abs/2201.10005} \end{APACrefURL}
\PrintBackRefs{\CurrentBib}

\bibitem [\protect \citeauthoryear {%
Sha%
\ \protect \BOthers {.}}{%
Sha%
\ \protect \BOthers {.}}{%
{\protect \APACyear {2021}}%
}]{%
sha2021fairness}
\APACinsertmetastar {%
sha2021fairness}%
\begin{APACrefauthors}%
Sha, L.%
, Rakovic, M.%
, Whitelock-Wainwright, A.%
, Carroll, D.%
, Yew, V\BPBI M.%
, Gasevic, D.%
\BCBL {}\ \BBA {} Chen, G.%
\end{APACrefauthors}%
\unskip\
\newblock
\APACrefYearMonthDay{2021}{}{}.
\newblock
{\BBOQ}\APACrefatitle {Assessing Algorithmic Fairness in Automatic Classifiers of Educational Forum Posts} {Assessing algorithmic fairness in automatic classifiers of educational forum posts}.{\BBCQ}
\newblock
\BIn{} I.~Roll, D.~McNamara, S.~Sosnovsky, R.~Luckin\BCBL {}\ \BBA {} V.~Dimitrova\ (\BEDS), \APACrefbtitle {Artificial Intelligence in Education} {Artificial intelligence in education}\ (\BVOL\ 12748, \BPGS\ 381--394).
\newblock
\APACaddressPublisher{}{Springer International Publishing}.
\newblock
\begin{APACrefDOI} \doi{10.1007/978-3-030-78292-4_31} \end{APACrefDOI}
\PrintBackRefs{\CurrentBib}

\bibitem [\protect \citeauthoryear {%
Torruella%
\ \BBA {} Capsada%
}{%
Torruella%
\ \BBA {} Capsada%
}{%
{\protect \APACyear {2013}}%
}]{%
torruella2013lexical_statistics}
\APACinsertmetastar {%
torruella2013lexical_statistics}%
\begin{APACrefauthors}%
Torruella, J.%
\BCBT {}\ \BBA {} Capsada, R.%
\end{APACrefauthors}%
\unskip\
\newblock
\APACrefYearMonthDay{2013}{}{}.
\newblock
{\BBOQ}\APACrefatitle {Lexical Statistics and Tipological Structures: A Measure of Lexical Richness} {Lexical statistics and tipological structures: A measure of lexical richness}.{\BBCQ}
\newblock
\APACjournalVolNumPages{Procedia - Social and Behavioral Sciences}{95}{}{447--454}.
\newblock
\begin{APACrefDOI} \doi{10.1016/j.sbspro.2013.10.668} \end{APACrefDOI}
\PrintBackRefs{\CurrentBib}

\bibitem [\protect \citeauthoryear {%
Winne%
}{%
Winne%
}{%
{\protect \APACyear {2017}}%
}]{%
winne2017learning_analytics}
\APACinsertmetastar {%
winne2017learning_analytics}%
\begin{APACrefauthors}%
Winne, P\BPBI H.%
\end{APACrefauthors}%
\unskip\
\newblock
\APACrefYearMonthDay{2017}{}{}.
\newblock
{\BBOQ}\APACrefatitle {Learning Analytics for Self-Regulated Learning} {Learning analytics for self-regulated learning}.{\BBCQ}
\newblock
\BIn{} \APACrefbtitle {Handbook of Learning Analytics} {Handbook of learning analytics}\ (\BPGS\ 531--566).
\newblock
\APACaddressPublisher{}{Society for Learning Analytics Research (SoLAR)}.
\PrintBackRefs{\CurrentBib}

\bibitem [\protect \citeauthoryear {%
Zhang%
\ \protect \BOthers {.}}{%
Zhang%
\ \protect \BOthers {.}}{%
{\protect \APACyear {2022}}%
}]{%
zhang2022using}
\APACinsertmetastar {%
zhang2022using}%
\begin{APACrefauthors}%
Zhang, J.%
, Andres, J\BPBI M\BPBI A\BPBI L.%
, Hutt, S.%
, Baker, R\BPBI S.%
, Ocumpaugh, J.%
, Nasiar, N.%
\BDBL {}others%
\end{APACrefauthors}%
\unskip\
\newblock
\APACrefYearMonthDay{2022}{}{}.
\newblock
{\BBOQ}\APACrefatitle {Using machine learning to detect SMART model cognitive operations in mathematical problem-solving process} {Using machine learning to detect smart model cognitive operations in mathematical problem-solving process}.{\BBCQ}
\newblock
\APACjournalVolNumPages{Journal of Educational Data Mining}{14}{3}{76--108}.
\PrintBackRefs{\CurrentBib}

\bibitem [\protect \citeauthoryear {%
Zhang%
, Borchers%
, Aleven%
\BCBL {}\ \BBA {} Baker%
}{%
Zhang%
\ \protect \BOthers {.}}{%
{\protect \APACyear {2024}}%
}]{%
zhang2024srl_llms}
\APACinsertmetastar {%
zhang2024srl_llms}%
\begin{APACrefauthors}%
Zhang, J.%
, Borchers, C.%
, Aleven, V.%
\BCBL {}\ \BBA {} Baker, R\BPBI S.%
\end{APACrefauthors}%
\unskip\
\newblock
\APACrefYearMonthDay{2024}{}{}.
\newblock
{\BBOQ}\APACrefatitle {Using large language models to detect self-regulated learning in think-aloud protocols} {Using large language models to detect self-regulated learning in think-aloud protocols}.{\BBCQ}
\newblock
\BIn{} \APACrefbtitle {Proceedings of the 17th International Conference on Educational Data Mining.} {Proceedings of the 17th international conference on educational data mining.}
\PrintBackRefs{\CurrentBib}

\end{thebibliography}

\newpage

\end{document}